\def\year{2022}\relax
\documentclass[letterpaper]{article} 
\usepackage{aaai22}  
\usepackage{times}  
\usepackage{helvet}  
\usepackage{courier}  
\usepackage[hyphens]{url}  
\usepackage{graphicx} 
\usepackage{natbib}  
\usepackage{caption} 
\DeclareCaptionStyle{ruled}{labelfont=normalfont,labelsep=colon,strut=off} 
\usepackage{algorithm}
\usepackage{algorithmic}

\usepackage{color}
\usepackage{enumitem}

\newcommand{\ieno}{\textit{i}.\textit{e}.}

\newcommand{\egno}{\textit{e}.\textit{g}.} 

\usepackage{newfloat}
\usepackage{listings}
\floatstyle{ruled}
\newfloat{listing}{tb}{lst}{}
\floatname{listing}{Listing}
\usepackage{epsfig}
\usepackage{graphicx}
\usepackage{amsmath}

\usepackage{amssymb}
\usepackage{makecell}

\usepackage{bm} 
\usepackage{multirow} 
\usepackage{hhline} 
\usepackage{array} 
\newcolumntype{M}[1]{>{\centering\arraybackslash}m{#1}} 
\usepackage[utf8]{inputenc}
\usepackage[english]{babel}

\usepackage{anyfontsize} 
\usepackage{mathtools} 

\newlength{\textfloatsepsave} 
\usepackage{varwidth}
\usepackage{lipsum,etoolbox}
\usepackage{graphics}

\title{Debiased Batch Normalization via Gaussian Process for Generalizable \\ Person Re-Identification}
\author{Paper ID: 3615}
\author{
Jiawei Liu\textsuperscript{\rm 1}\equalcontrib, 
Zhipeng Huang\textsuperscript{\rm 1}\equalcontrib, 
Liang Li\textsuperscript{\rm 2}, 
Kecheng Zheng\textsuperscript{\rm 1}, 
Zheng-Jun Zha\textsuperscript{\rm 1}\thanks{Corresponding Author}}
\affiliations {
    \textsuperscript{\rm 1} University of Science and Technology of China \\
    \textsuperscript{\rm 2} Institute of Computing Technology, Chinese Academy of Sciences \\
    \{hzp1104,zkcys001\}@mail.ustc.edu.cn, \{jwliu6,zhazj\}@ustc.edu.cn, liang.li@ict.ac.cn
}
\usepackage{bibentry}

\begin{document}

\maketitle

\begin{abstract}
Generalizable person re-identification aims to learn a model with only several labeled source domains that can perform well on unseen domains. Without access to the unseen domain, the feature statistics of the batch normalization (BN) layer learned from a limited number of source domains is doubtlessly biased for unseen domain. This would mislead the feature representation learning for unseen domain and deteriorate the generalizaiton ability of the model. In this paper, we propose a novel Debiased Batch Normalization via Gaussian Process approach (GDNorm) for generalizable person re-identification, which models the feature statistic estimation from BN layers as a dynamically self-refining Gaussian process to alleviate the bias to unseen domain for improving the generalization. Specifically, we establish a lightweight model with multiple set of domain-specific BN layers to capture the discriminability of individual source domain, and learn the corresponding parameters of the domain-specific BN layers. These parameters of different source domains are employed to deduce a Gaussian process. We randomly sample several paths from this Gaussian process served as the BN estimations of potential new domains outside of existing source domains, which can further optimize these learned parameters from source domains, and estimate more accurate Gaussian process by them in return, tending to real data distribution. Even without a large number of source domains, GDNorm can still provide debiased BN estimation by using the mean path of the Gaussian process, while maintaining low computational cost during testing. Extensive experiments demonstrate that our GDNorm effectively improves the generalization ability of the model on unseen domain.

	
\end{abstract}

\section{Introduction}

Person re-identification (Re-ID)~\cite{ye2021deep,luo2019strong,liu2016multi,liu2019adaptive} aims to identify a person-of-interest across non-overlapping camera networks under different times and locations.It has drawn extensive research attention in recent years, due to its significant role in various practice scenarios, such as object tracking, activity analysis and smart retail \textit{etc}~\cite{wang2013intelligent,caba2015activitynet,martini2020open}. Along with the success of deep learning technique, large amount of sophisticated person Re-ID methods have been proposed and achieved promising performances in a fully supervised manner, which focus on handling the challenge of cluttered background, partial occlusion, viewpoint and pose variations \textit{etc}. 

These approaches perform well when trained and tested on separated splits of the same domain (dataset). However, when applying these trained models to previously unseen domains, they suffer from significant performance degradation due to style discrepancies across domains. Thus, it is imperative to learn a Re-ID model that has a strong generalization ability to unseen domain. To this end, unsupervised domain adaptation (UDA) has been introduced in person Re-ID. UDA approaches employ unlabeled samples from target domain to adapt the pre-trained model on the labeled source domain to target domain \cite{bak2018domain}. Although UDA approaches are more practical than the full supervised approaches, they still require data collection and model update for the target domain, resulting in additional computational cost. Moreover, samples from a target domain are usually not unavailable in advance. 

Compared to UDA, domain generalization (DG) is a promising solution for real-world applications. Domain generalization aims at learning models that are generalizable to unseen target domain, without requiring access to the data and annotations from target domain as well as model updating, \textit{i.e.}, fine-tune or adaptation. Most existing DG methods~\cite{chattopadhyay2020learning,huang2020self} including data manipulation, representation learning and learning strategy, assume that the source and target domains share the same label space with a fixed of classes, which are designed for a classification task. In contrast, domain generalization for person Re-ID is an open-set retrieval task, having different and variable number of identities between source and target domains. Therefore, it is difficult to achieve satisfying generalization capability when the existing DG approaches are directly applied to person Re-ID.

Existing generalizable person Re-ID approaches can be divided into two categories: single model methods~\cite{zhao2020learning, choi2020meta,song2019generalizable} and ensemble learning based methods~\cite{dai2021generalizable,seo2020learning,zhuang2020rethinking,mancini2018best}. The former collects all source domain data and trains a single model on them to extract the shared domain-invariant representations. They mainly design customized loss functions (\textit{e.g.}, Maximum Mean Discrepancy regularization to align the distributions among different domain), specific architectures (\textit{e.g.} integrating Batch Normalization and Instance Normalization in the models to alleviate the domain discrepancy due to appearance style variations), or ad-hoc training policies (\textit{e.g.}, model-agnostic meta learning to mimic real train-test domain shift). However, these methods discard the
specific characteristics of individual source domains and neglect the inherent relevance of unseen target domain with respect to source domains, result in unsatisfying generalization capability. Such domain-specific characteristics can provide adequately discriminative and meaningful information and infer the distribution of unseen target domain, thus greatly improving the model’s generalizability.

Ensemble learning based methods attempt to train domain-specific models for each source domain by exploiting domain specific characteristics, and combine multiple domain-specific models (\egno, branches, classifiers or experts) to enhance the generalization ability. These methods assume that any sample can be considered as a comprehensive sample from multi-source domains, such that the overall prediction result can be seen as the superposition of multi-domain networks. Nevertheless, these works have two main drawbacks: (1) The number of source domains in DG person Re-ID task is extremely limited, directly ensembling them results in a high bias to the underlying hyper-distribution (real data distribution) on domains. The biased issue injury the generalization ability of the model; (2) Ensemble learning significantly increases the computational cost due to training and testing multiple neural networks. When there exists a large amount of source domains, the computational cost of the ensemble learning based method is unacceptable. Thus, how to efficiently utilize the limited number of source domains to improve the generalization is the major issue for generalizable person re-identification.


In this work, we propose a novel Debiased Batch Normalization via Gaussian Process approach (GDNorm) for DG person Re-ID, which models the feature statistic estimation from BN layers as a dynamically  self-refining Gaussian process to alleviate the bias issue on unseen domain for improving the generalization. GDNorm establishes a lightweight model with multiple set of domain-specific BN layers for source domains to learn the corresponding population statistics and rescaling parameters of the domain-specific BN layers. The parameters of the domain-specific BN layers from different source domains are assumed to fit a Gaussian process, and act as an approximate distribution of the domain-specific model for each source domain. We randomly sample paths (the parameters of the BN layers) from the deduced Gaussian process as the BN estimations of potential new domains, which can further optimize these parameters, and estimate more accurate Gaussian process by them in return, tending to the underlying hyper-distribution (real data distribution) on domains. Such self-refining  mechanism guarantees that even with a limited number of source domains, GDNorm can still provide debiased BN estimation on unseen domain. In the testing phase, GDNorm directly exploits the mean path of the Gaussian process as the optimal BN estimating for target domain, greatly reducing the computational cost as compared to ensemble learning. Extensive experiments have demonstrated the effectiveness of the proposed GDNorm when testing on the unseen domain.

The main contributions of this paper are three-fold: (1) We propose a novel Debiased Batch Normalization via Gaussian Process approach (GDNorm) for generalizable person Re-ID. (2) We design a novel debiased Batch Normalization, which models the feature statistic estimation of BN as a dynamically self-refining Gaussian process, alleviating the bias issue of the BN estimation on unseen domain. (3) We propose a lightweight model for generalizable person Re-ID, which exploits the mean path of the Gaussian process as the optimal debiased BN parameters, without requiring extra computational cost as compared to ensemble learning.


\section{Related Work}

\subsection{DG Person Re-Identification}

Existing DG person Re-ID methods \cite{zhao2020learning, choi2020meta,song2019generalizable,lin2020multi,chen2020dual, jin2020style, tamura2019augmented} can be divided into two categories: single model methods and ensemble learning based methods. For the single model methods, Chen \textit{et al.} \cite{chen2020dual} proposed a Dual Distribution Alignment Network with dual-level constraints, \textit{i.e.}, a domainwise adversarial feature learning and an identity-wise similarity enhancement, which maps pedestrian images into a domain-invariant feature space. Choi \textit{et al.} \cite{choi2020meta} designed learnable batch-instance normalization layers, which prevents the model from overfitting to the source domains by the simulation of unsuccessful generalization scenarios in meta-learning pipeline. Lin \textit{et al.} \cite{lin2020domain} present an episodic learning scheme, which advances meta-learning algorithm to exploit the labeled data from source domain and learns domain-invariant features without observing target domain data. 
The ensemble learning based methods will be elaborated in the next subsection.

\begin{figure*}[!t]
	\centering
	\includegraphics[width=0.99\textwidth]{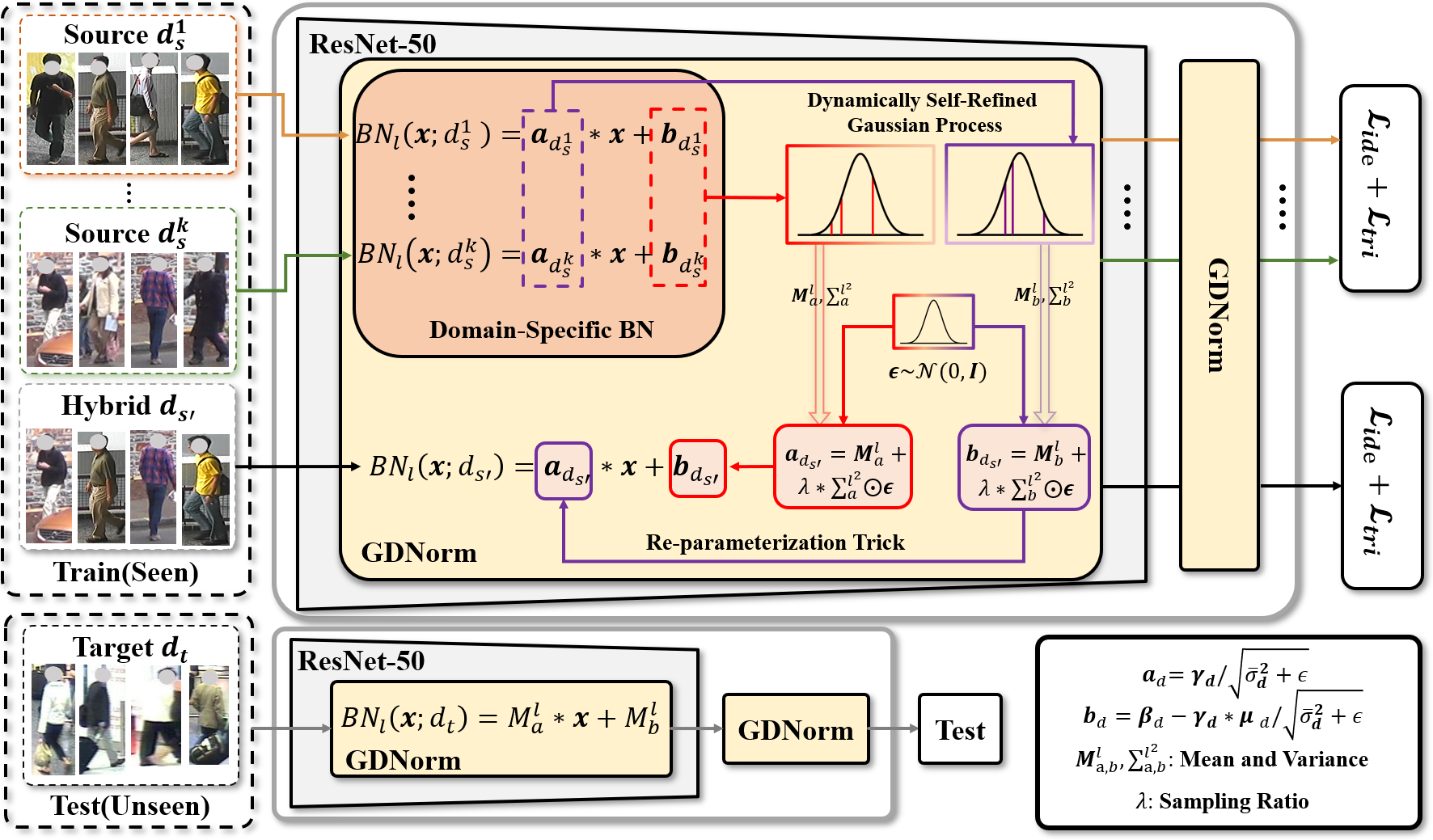}
	\caption{The overall architecture of the proposed GDNorm. It contains a lightweight model with multiple set of domain-specific BN layers for capturing the specific feature statistics information, and a dynamically self-refining Gaussian process estimation for learning debiased BN parameters on unseen domain.}
	\label{fig:framework}
\end{figure*}




\subsection{Ensemble learning based DG}
Ensemble learning based methods ensemble multiple domain specific models like experts or classifiers to enhance the generalization ability of the overall network, which are applied to general DG and GD person Re-ID tasks. Mancini \textit{et al.}~\cite{mancini2018best} proposed to fuse the predictions from different domain-specific classifiers with learnable weight and a domain predictor. D’Innocente \textit{et al.}~\cite{d2018domain} proposed domain-specific layers of different source domains and learned the linear aggregation of these layers to represent a test sample. Zhou \textit{et al.}~\cite{zhou2020domain} proposed a domain adaptive ensemble learning method, which improves the generalization of a multi-expert network by explicitly training the ensemble to solve the target task. 
Dai \textit{et al.}~\cite{dai2021generalizable} proposed a method called the relevance-aware mixture of experts (RaMoE), by using an effective voting-based mixture mechanism to dynamically leverage source domains’ diverse characteristics to improve the generalization.

\subsection{Normalization in DA and DG}
Normalization techniques in deep neural networks are designed for regularizing trained models and improving their generalization performance. Recently, several methods on domain adaptation (DA) and domain generalization (DG) discovered the relationship between domain gap and normalization operation. For example, Jin \textit{et al.}~\cite{jin2021style} proposed a Style Normalization and Restitution module, which utilizes the Instance Normalization (IN) layers to filter out style variations and compensates for the identity-relevant features discarded by IN layers. Seo \textit{et al.}~\cite{seo2020learning} proposed to leverage instance normalization to optimize the trade-off between cross-category variance and domain invariance, which is desirable for domain generalization in unseen domains. Zhuang \textit{et al.}~\cite{zhuang2020rethinking} proposed camera-based batch normalization (CBN) to force the images of all cameras to fall onto the same subspace and to shrink the distribution gap between any camera pair.


\section{Method}
\subsection{Bias in Generalizable Person Re-ID}
For DG person Re-ID, we assume that we have $K$ source domains (datasets) $\mathcal{D_S} = \{d_s^1, d_s^2,..., d_s^K\}$ in the training stage. Each domain $d_s^k= \{(\textbf{x}_i^k,y_i^k)\}_{i=1}^{N_k} $ has its own label space (person IDs among different datasets are non-overlapping), where $N_k$ is the total number of pedestrian images in the source domain $d_s^k$, each image $\textbf{x}_i^k$ is  associated with an identity label $y_i^k$. During the testing stage, the trained model is frozen and directly deployed to a previously unseen dataset (target domain $d_t$) without additional model updating. More formally, we denote $\mathcal{X}$ as the input pedestrian images space, and denote $\mathcal{Y}$  as the label space. A domain (dataset) in DG person Re-ID can be denotes as the aggregation of data sampled from a specific joint distribution $d^k={(\textbf{x}_i,y_i)}_{i=1}^{N_k} \sim P_{\mathrm{xy}}^{d^k}$. Their joint distributions of source and target domains are completely different. DG person Re-ID aims to learn a generalizable model $\overline{P}(Y|X)$ which achieves the smallest prediction error on unknown target domain without fine-tuning:
\begin{equation}
\min _{\overline{P}} \mathbb{E}_{(\mathbf{x}, y) \sim P_{\mathrm{xy}}^{d_t}}[\ell(\overline{P}(Y|X=\mathbf{x}), y)]
\end{equation}
where $\ell$ denotes a loss function on $\mathcal{Y}$. Due to the fact that the target domain is totally unknown (not even unsupervised data), it is required to measure the average risk over all possible target domains. Assuming both source and target domain distributions follow an underlying hyper-distribution $\mathcal{H}$: $P_{\mathrm{xy}} \sim \mathcal{H}$. Then the target prediction error (risk) is reformulated as the following:
\begin{equation}
    \mathcal{E}:=\mathbb{E}_{P_{\mathrm{xy}} \sim \mathcal{H}} \mathbb{E}_{(\mathbf{x}, y) \sim P_{\mathrm{xy}}} \left[\ell\left(\overline{P}(Y|X=\textbf{x}), y\right)\right]
    \label{eq:average risk}
\end{equation}

Existing ensemble learning based DG methods~\cite{mancini2018best,segu2020batch,d2018domain,zhou2020domain} simply regard the hyper-distribution $\mathcal{H}$ as a static and discrete distribution with respect to the limited source domains: $ \{P_{\mathcal{H}}(P_{\mathrm{xy}}=P_{\mathrm{xy}}^{d_s^k}) = \pi^{d_s^k}\}_{k=1}^K$. Under this approximate hypothesis, they exploit the relationship between multiple source domains by designing specific network architecture and training strategies to improve generalization. They assume that any sample can be regarded as an integrated sample of the multiple source domains, such that the overall prediction result can be seen as the superposition of the multiple domain networks. Thus, the approximate optimal solution for Eq.~\ref{eq:average risk} is then computed as the following:
\begin{equation}
    \overline{P}(Y|X)=\sum_{k=1}^K \pi^{d_s^k} \overline{P}(Y|X, P_{\mathrm{xy}}^{d_s^k})
    \label{eq:ensemble}
\end{equation}
where $\overline{P}(Y|X,P_{\mathrm{xy}}^{d_s^k})$ represents the $k$-th source domain-specific model, and $\pi^k$ denotes the mixture ratio. Nevertheless, these methods employ such simple discrete, static distribution to represent the complex hyper-distribution $\mathcal{H}$, leading to high bias. First, the discrete distribution considers all possible domains under $\mathcal{H}$ as a mixture of a limited number of  source domains. Only using a small number of domains could definitely not express the complex unknown $\mathcal{H}$. Second, the static distribution is not able to reduce and adjust the deviation to the real hyper-distribution $\mathcal{H}$. In addition, ensemble model leads to a heavy increasing of computational cost, due to maintaining multiple domain specific networks in the testing stage.

%
Unlike these methods, we rethink the hyper-distribution $\mathcal{H}$ as a dynamically self-refining Gaussian process. In theory, a linear combination of multiple Gaussian distributions or Gaussian process can fit any type of distribution~\cite{bond2001gmm,yu2019adaptive}, which is a better choice to serve as the hyper-distribution $\mathcal{H}$. Based on this, we design a self-refining mechanism to refine and debias the estimated gaussian process, towards obtaining the ideal unbiased optimal solution of Eq.~\ref{eq:average risk}:
\begin{equation}
    \begin{aligned}
    \overline{P}_{unbiased}(Y|X)  & = \int_{P_{\mathrm{xy}}\sim H} P(P_{\mathrm{xy}}|X) \overline{P}(Y|X, P_{\mathrm{xy}}) \\
    & =\mathbb{E}_{\mathcal{H}|\textbf{x}}\overline{P}(Y|X, P_{\mathrm{xy}})\\
    \end{aligned}
    \label{eq:total probability}
\end{equation}
Specifically, we propose a novel Debiased Batch Normalization via Gaussian Process approach (GDNorm) for generalizable person Re-ID, as illustrated in Fig.~\ref{fig:framework}. GDNorm establishes a lightweight model $\overline{P}(Y|X, P_{\mathrm{xy}})$ with different set of domain-specific BN layers, which is based on the popular ResNet-50 model \cite{he2016deep}. The expected domain specific models for each source domain share all the training parameters of the lightweight model apart from the learned parameters $\boldsymbol{\theta}_{d_s}$ from the domain-specific BN layers. GDNorm utilizes these parameters to serve as the distribution of the domain-specific model, $\overline{P}(Y|X=x, P_{\mathrm{xy}}^{d_s})$  \textit{s.t.} $P_{\mathrm{xy}}^{d_s}\sim\mathcal{H}$, which follows a Gaussian process. It then randomly samples paths (the parameters of domains-specific BN layers) from this Gaussian process to obtain the BN estimations of potential new domains outside of the training source domains. The sampled paths can in return, facilitate the Gaussian process close to (debias to) the distribution of the hyper-distribution by end-to-end optimizing. In the testing phase, unlike the previous ensemble learning based methods, GDNorm directly employs the mean path of the learned debiased Gaussian process as optimal batch normalization layers' parameters on unseen target domain for model inference, without requiring extra computational cost.




\subsection{Domain-Specific Batch Normalization}
BN is a widely-used training technique in deep networks, which normalizes the activations from each domain to the same reference distribution by domain-specific normalization statistics \cite{chang2019domain}. To capture the specific characteristics of individual domain, a lightweight model with different set of domain-specific BN layers is designed, which can capture the source domain distribution $\tiny P_{\mathrm{xy}}^{d_s}$.

Concretely, given a batch size of pedestrian images from a certain source domain $d_s$ at the training stage, a domain-specific BN layer normalizes the activations by matching their first and second order moments $(\boldsymbol{\mu}_{d_s}, \boldsymbol{\sigma}_{d_s}^2)$ to a reference Gaussian distribution with zero mean and unitary variance $\mathcal{N}(0, 1)$. And then transforms channel-wise whitened activations by using rescaling parameters. The formulation of a BN layer is defined as the following:
\begin{equation}
	\begin{aligned}
		BN(\boldsymbol{x}; d_s)=\boldsymbol{\gamma}_{d_s}\frac{\boldsymbol{x}-\boldsymbol{\mu}_{d_s}}{\sqrt{\boldsymbol{\sigma}_{d_s}^2+\epsilon}} + \boldsymbol{\beta}_{d_s}
	\end{aligned}
\end{equation}
where $\boldsymbol{x}$ refers to an input activation from the domain $d_s$, $\epsilon>0$ is a small constant to avoid numerical problem. $\boldsymbol{\gamma}_{d_s}$ and $\boldsymbol{\beta}_{d_s}$ are the learnable rescaling parameters, which are optimized by samples from the specific source domain. $\boldsymbol{\mu}_{d_s}$ and $\boldsymbol{\sigma}_{d_s}^2$ are the domain-specific batch statistics for the domain $d_s$, which are calculated as the following:
\begin{equation}
	\begin{aligned}
		\boldsymbol{\mu}_{d_s} &= \frac{\sum_{b, i, j} \boldsymbol{x}[b, i, j]}{B \cdot H \cdot W} \\
		\boldsymbol{\sigma}_{d_s}^2 &= \frac{\sum_{b, i, j}\left(\boldsymbol{x}[b, i, j]-\boldsymbol{\mu}_{d_s}\right)^{2}}{B \cdot H \cdot W}
	\end{aligned}
\end{equation}
where $B$ denotes the batch size, $H$ and $W$ is the height and width size of the input activations. The domain-specific batch statistics are updated by using moving average strategy as the following:
\begin{equation}
	\begin{aligned}
		\overline{\boldsymbol{\mu}}_{d_s} &= m \overline{\boldsymbol{\mu}}_{d_s} + (1-m) \boldsymbol{\mu}_{d_s} \\
		\overline{\boldsymbol{\sigma}}_{d_s}^2 &= m \overline{\boldsymbol{\sigma}}_{d_s}^2 + (1-m) \boldsymbol{\sigma}_{d_s}^2
	\end{aligned}
\end{equation}
where $	\overline{\boldsymbol{\mu}}_{d_s}$, $\overline{\boldsymbol{\sigma}}_{d_s}^2$ reflect  domain-specific population statistics of this source domain across all batch data. The expected domain-specific model with the parameters $\boldsymbol{\theta}_{d_s}$ of the domain-specific BN layers for each source domain is then learned as the following:
\begin{equation}
\begin{aligned}
    &\boldsymbol{\theta}_{d_s} = \{(\boldsymbol{\gamma}_{d_s}^l, \boldsymbol{\beta}_{d_s}^l, \overline{\boldsymbol{\mu}}_{d_s}^l, \overline{\boldsymbol{\sigma}}_{d_s}^{l^2})\}_{l=1}^{L} \\
    &\overline{P}(Y|X, P_{xy}^{d_s}) = \overline{P}(Y|X, \boldsymbol{\theta}_{d_s}) \\
\end{aligned}
\end{equation}
where $L$ is the number of batch normalization layers in our lightweight model for each source domain. 






\subsection{Debias Gaussian Process}
The learned parameters $\boldsymbol{\theta}_{d_s}$ of the domain-specific BN layers contain specific feature statistics information of the corresponding domain ${d_s}$. Therefore, the distribution of the parameters $\boldsymbol{\theta}_{d_s}$ can be represented as the hyper-distribution $\mathcal{H}$ on domains. We assume that the distributions of the parameters $\boldsymbol{\theta_{d_s}}$ in the model follow a Gaussian process (A collection of Gaussian distributions along different depth position BN layers in the model). GDNorm deduces a dynamical Gaussian process from the learned parameters $\boldsymbol{\theta_{d_s}}$ of the domain-specific BN layers to fit the hyper-distribution $\mathcal{H}$. To reduce the bias between the Gaussian process and the hyper-distribution, it designs a self-refining mechanism to randomly sample several paths from the estimated Gaussian process to serve as the BN estimations of potential new domains outside of source domains, which can further optimize these parameters, and infer more accurate Gaussian process by them in return, tending to real hyper-distribution.



%
For simplicity, we reformulate the BN operation into a linear transformation as the following:
\begin{equation}
    \begin{aligned}
    BN(\boldsymbol{x}; d_s) &=\boldsymbol{a}_{d_s} \boldsymbol{x} + \boldsymbol{b}_{d_s} \\
    \boldsymbol{a}_{d_s} &= \frac{\boldsymbol{\gamma}_{d_s}}{\sqrt{\overline{\boldsymbol{\sigma}}_{d_s}^2+\epsilon}} \\
    \boldsymbol{b}_{d_s} &=  \boldsymbol{\beta}_{d_s} - \boldsymbol{\gamma}_{d_s} \frac{\overline{\boldsymbol{\mu}}_{d_s}}{\sqrt{\overline{\boldsymbol{\sigma}}_{d_s}^2+\epsilon}}
    \end{aligned}
\end{equation}
GDNorm calculates the mean $\boldsymbol{M}$ and variance $\boldsymbol{\Sigma}^2$ of the learned parameters in each domain-specific BN layer for all source domains to obtain a collection of Gaussian distributions along different depth position BN layers in the model.
\begin{equation}
	\begin{aligned}
		\boldsymbol{M}(\boldsymbol{a}^l) &= \frac{\sum_{k=1}^K \boldsymbol{a}_{d_s^k}^l}{K} \quad \boldsymbol{M}(\boldsymbol{b}^l) = \frac{\sum_{k=1}^K \boldsymbol{b}_{d_s^k}^l}{K} \\
		\boldsymbol{\Sigma}^2(\boldsymbol{a}^l) &= \frac{\sum_{k=1}^K  (\boldsymbol{a}_{d_s^k}^l - \boldsymbol{M}(a^l))^{2}}{K} \\
		\boldsymbol{\Sigma}^2(\boldsymbol{b}^l) &= \frac{\sum_{k=1}^K  (\boldsymbol{b}_{d_s^k}^l - \boldsymbol{M}(b^l))^{2}}{K}
	\end{aligned}
\end{equation}
These Gaussian distributions constitute a Gaussian process whose random variables are indexed by the different depth position: $GP = \{(\boldsymbol{M}^l, \boldsymbol{\Sigma  }^{l^2})\}_{l=1}^L$. 

To refine the estimated Gaussian process unbiased to the real hyper-distribution, GDNorm randomly samples several paths from the Gaussian process as the BN estimations of potential domains for further training and updating these learned parameters. It then employs a reparameterization trick~\cite{blundell2015weight} to bypass the discrete sampling process, which allows end-to-end optimizing:
\begin{equation}
\boldsymbol{a}_{s'}^l,\boldsymbol{b}_{s'}^l = \boldsymbol{M}^l+\lambda \boldsymbol{\Sigma}^{l^2} \odot \boldsymbol{\epsilon}
\label{eq:sample}
\end{equation}
where $\{\boldsymbol{a}_{s'}^l,\boldsymbol{b}_{s'}^l\}_{l=1}^L$ is a randomly sample path from the Gaussian process, corresponding to a potential new domain $d_{s'}$. $\boldsymbol{\epsilon} \sim \mathcal{N}(\mathbf{0}, \boldsymbol{I})$ and $\lambda$ is the sampling ratio. The randomly sampled path is used as the BN parameters for the potential new domain. A batch of images sampled equally from all existing source domains as the samples of the potential new domain to train the model, which is supervised by a identity loss and a triplet loss~\cite{schroff2015facenet}. The parameters of the domain-specific BN layer for the existing source domains will be updated by this optimizing process, which can be used to deduce more accurate Gaussian process by them in return, tending to real hyper-distribution. In the testing phase, GDNorm directly utilizes the mean path of the final debiased Gaussian process $\{(\mathcal{M}(\boldsymbol{a}^l), \mathcal{M}(\boldsymbol{b}^l))\}_{l=1}^L$ to serve as every BN layers in the model, runs the trained model only once for inference, without requiring extra computational cost as compared to ensemble learning.


\section{Experiments}

\begin{table*}[tb]
	\caption{Performance (\%) comparison with the state-of-the-art methods under Protocol-1.}  
	\centering
	\vspace{-0.20cm}
	\resizebox{\linewidth}{!}{
		\begin{tabular}{c|c|c||ccc||ccc||ccc||ccc||ccc}
			\Xhline{3\arrayrulewidth}
			\multirow{2}{*}{Method} & \multirow{2}{*}{Type} & \multirow{2}{*}{Source} & \multicolumn{3}{c||}{Target: VIPeR (V)} & \multicolumn{3}{c||}{Target: PRID (P)} & \multicolumn{3}{c||}{Target: GRID (G)} & \multicolumn{3}{c||}{Target: iLIDS (I)} & \multicolumn{3}{c}{Mean of V,P,G,I}\\ \cline{4-18} 
			&&& R1 & R5  & \emph{m}AP & R1 & R5  & \emph{m}AP & R1 & R5  & \emph{m}AP & R1 & R5 & \emph{m}AP & R1 & R5 & \emph{m}AP \\ \hline
            ImpTrpLoss~\cite{cheng2016person}               & S & Target  & 47.8 & 74.7  & -    & 22.0 & -     & -    & -    & -     & -    & 60.4 & 82.7  & -    & - & - & -\\
            JLML~\cite{li2017person}                        & S & Target  & 50.2 & 74.2  & -    & -    & -     & -    & 37.5 & 61.4  & -    & -    & -     & -    & - & - & -\\
            SSM~\cite{bai2017scalable}                      & S & Target  & 53.7 & -     & -    & -    & -     & -    & 27.2 & -     & -    & -    & -     & -    & - & - & -\\ \hline
            TJ-AIDL~\cite{wang2018transferable}             & UDA & M     & 38.5 & -     & -    & 26.8 & -     & -    & -    & -     & -    & -    & -     & -    & - & - & -\\
            MMFA~\cite{lin2018multi}                        & UDA & M     & 39.1 & -     & -    & 35.1 & -     & -    & -    & -     & -    & -    & -     & -    & - & - & -\\
            UDML~\cite{peng2016unsupervised}                & UDA & Comb1 & 31.5 & -     & -    & 24.2 & -     & -    & -    & -     & -    & 49.3 & -     & -    & - & - & -\\
            SyRI~\cite{bak2018domain}                       & UDA & Comb2 & 43.0 & -     & -    & 43.0 & -     & -    & -    & -     & -    & 56.5 & -     & -    & - & - & -\\ \hline
            CDEL~\cite{lin2020domain}                       & DG & MS     & 38.5 &  -    &   -  & 57.6 &   -   &   -  & 33.0 &   -   &   -  & 62.3 &   -   &   -   & 47.9 & - & -\\
            DIMN~\cite{song2019generalizable}               & DG & MS     & 51.2 & 70.2  & 60.1 & 39.2 & 67.0  & 52.0 & 29.3 & 53.3  & 41.1 & 70.2 & 89.7  & 78.4  & 47.5 & 70.1 & 57.9 \\
            AugMining~\cite{tamura2019augmented}            & DG & MS     & 49.8 & 70.8  &  -   & 34.3 & 56.2  &  -   & 46.6 & 67.5  &  -   & 76.3 & 93.0  & -     & 51.8 & 71.9 & - \\
            DualNorm~\cite{jia2019frustratingly}            & DG & MS     & 53.9 & 62.5  & 58.0 & 60.4 & 73.6  & 64.9 & 41.4 & 47.4  & 45.7 & 74.8 & 82.0  & 78.5  & 57.6 & 66.4 & 61.8\\
            DDAN~\cite{chen2020dual}                        & DG & MS     & 56.5 & 65.6  & 60.8 & 62.9 & 74.2  & 67.5 & 46.2 & 55.4  & 50.9 & 78.0 & 85.7  & 81.2  & 60.9 & 70.2 & 65.1\\
            RaMoE~\cite{dai2021generalizable}               & DG & MS     & 56.6 & -     & 64.6 & 57.7 & -     & 67.3 & 46.8 & -     & 54.2 & \textbf{85.0} & -     & \textbf{90.2}  & 61.5 & - & 69.1\\
            DIR-ReID~\cite{zhang2021learning}               & DG & MS     & 58.3 & 66.9  & 62.9 & 71.1 & 82.4  & 75.6 & 47.8 & 51.1  & 52.1 & 74.4 & 83.1  & 78.6  & 62.9 & 70.8 & 67.3\\
            \textbf{GDNorm (Ours)}                          & DG & MS & \textbf{66.1} & \textbf{83.5} & \textbf{74.1} & \textbf{72.6} & \textbf{89.3} & \textbf{79.9} & \textbf{55.4} & \textbf{73.8} & \textbf{63.8} & 81.3 & \textbf{94.0} & 87.2  & \textbf{68.9} & \textbf{85.2} & \textbf{76.3} \\ \Xhline{3\arrayrulewidth}
		\end{tabular}
	}
	 \vspace{-0.30cm}
	\label{table:result1}
\end{table*}

\begin{table}[!t]
	\centering
	\caption{Different evaluation settings of Protocol-1 and Protocol-2.}
	\vspace{-0.20cm}
	\resizebox{\linewidth}{!}{
	\begin{tabular}{c|c|c}
		\Xhline{3\arrayrulewidth}
		Setting                       & Training Data & Testing Data \\
		\hline
		Protocol-1                     & M+D+C2+C3+CS & PRID, GRID, VIPeR, iLIDs \\ \hline
		\multirow{4}{*}{Protocol-2} & D+C3+MT & M \\ \cline{2-3}
		& M++C3+MT & D \\ \cline{2-3}
		& M+D+MT & C3 \\ \cline{2-3}
		& M+D+C3 & MT \\
		\Xhline{3\arrayrulewidth}
	\end{tabular}%
	}
	\vspace{-0.10cm}
	\label{table:Protocol}%
\end{table}%


\begin{table}[!t]
	\caption{Performance (\%) comparison with the state-of-the-art methods under Protocol-2.}  
	\vspace{-0.20cm}
	\centering
	\resizebox{\linewidth}{!}{
	\begin{tabular}{c|cccc}
		\Xhline{3\arrayrulewidth}
		D+MT+C3$\rightarrow$M         & R1 & R5 &  R10 & \emph{m}AP  \\ \hline
		$QAConv_{50}$*~\cite{liao2019interpretable}               & 65.7 &  -   &   -   & 35.6  \\\
		$M^3L$(IBN-Net50)~\cite{zhao2020learning}            & 75.9 &  -   &   -   & 50.2  \\
		RaMoE~\cite{dai2021generalizable} & 82.0 & 91.4 & 94.4 & 56.5 \\
		\textbf{GDNorm (Ours)}      & \textbf{86.5} & \textbf{95.2} & \textbf{96.9} & \textbf{68.2}   \\ \hline
		M+MT+C3$\rightarrow$D         & R1 & R5 &  R10 & \emph{m}AP  \\ \hline
		$QAConv_{50}$*~\cite{liao2019interpretable}               & 66.1 &  -   &   -   & 47.1  \\
		$M^3L$(IBN-Net50)~\cite{zhao2020learning}            & 69.2 &  -   &   -   & 51.1  \\
		RaMoE~\cite{dai2021generalizable} & 73.6 & 85.3 & 88.4 & 56.9 \\
		\textbf{GDNorm (Ours)}      & \textbf{78.2} & \textbf{87.8} & \textbf{90.4}  & \textbf{63.8}  \\ \hline
		D+MT+M$\rightarrow$C3         & R1 & R5 &  R10& \emph{m}AP  \\ \hline
		$QAConv_{50}$*~\cite{liao2019interpretable}               & 23.5 &  -   &   -   & 21.0  \\
		$M^3L$(IBN-Net50)~\cite{zhao2020learning}            & 33.1 &  -   &   -   & 32.1  \\
		RaMoE~\cite{dai2021generalizable} & 34.6 & 54.3 & 64.6 & 33.5 \\
		\textbf{GDNorm (Ours)}      & \textbf{48.6} & \textbf{70.3} & \textbf{79.3}  & \textbf{47.9}  \\ \hline
		D+M+C3$\rightarrow$MT         & R1 & R5 &  R10 & \emph{m}AP  \\ \hline
		$QAConv_{50}$*~\cite{liao2019interpretable}               & 24.3 &  -   &   -   & 7.5   \\
		$M^3L$(IBN-Net50)~\cite{zhao2020learning}            & 33.0 &  -   &   -   & 12.9  \\
		RaMoE~\cite{dai2021generalizable} & 34.1 & 46.0 & 51.8 & 13.5 \\
		\textbf{GDNorm (Ours)}      & \textbf{48.1} & \textbf{63.1} & \textbf{68.5} & \textbf{20.4}  \\
		\Xhline{3\arrayrulewidth}
	\end{tabular}
	}
	\vspace{-0.50cm}
	\label{table:result2}
\end{table}

\begin{table}[tb]
	\caption{Ablation studies on the effectiveness of each component of GDNorm under Protocol-2.}  
	\vspace{-0.20cm}
	\centering
	\begin{tabular}{cc|cccc}
		\Xhline{3\arrayrulewidth}
		\multirow{2}{*}{E}  & \multirow{2}{*}{S} & \multicolumn{4}{c}{D+MT+C3$\rightarrow$M} \\ \cline{3-6}
		&& R1 & R5 & R10 & mAP \\ \hline
		$\times$ & $\times$ & 78.5 & 91.4 & 94.5 & 55.2 \\
		$\checkmark$ & $\times$ & 83.2 & 92.9 & 95.3 & 62.5 \\
		$\checkmark$ & $\checkmark$ & \textbf{86.5} & \textbf{95.2} & \textbf{96.9} & \textbf{68.2} \\
		\Xhline{3\arrayrulewidth}
	\end{tabular}
	\vspace{-0.20cm}
	\label{table:components}
\end{table}

\subsection{Datasets and Evaluation Settings}
\textbf{Protocol-1:} Following the previous methods~\cite{song2019generalizable,jia2019frustratingly,tamura2019augmented}, we employ the existing Re-ID benchmarks to evaluate the Re-ID model's generalization ability, where the existing large-scale Re-ID datasets are viewed as multiple source domains, and the small-scale Re-ID datasets are used as unseen target domains. As shown in Tab.~\ref{table:Protocol}, source domains include CUHK02~\cite{li2013locally}, CUHK03~\cite{li2014deepreid}, Market-1501~\cite{zheng2015scalable}, DukeMTMC~\cite{zheng2017unlabeled} and CUHK-SYSU~\cite{xiao2017joint}. Target domains contain VIPeR~\cite{gray2008viewpoint}, PRID~\cite{hirzer2011person}, GRID~\cite{loy2009multi} and iLIDS~\cite{zheng2009associating}. All training sets and testing sets in the source domains are used for model training. The four small-scale Re-ID datasets are tested respectively, where the final performances are obtained by the average of 10 repeated random splits of testing sets.

\textbf{Protocol-2:} Considering that the image quality of the small-scale Re-ID datasets is quite poor, the performances on these datasets can not precisely reflect the generalization ability of a model in real scenarios. The previous methods~\cite{zhao2020learning,dai2021generalizable} thus set a new protocol (\ieno, leave-one-out setting) for four large-scale Re-ID datasets. Specifically, four large-scale Re-ID datasets (Market-1501~\cite{zheng2015scalable}, DukeMTMC~\cite{zheng2017unlabeled}, CUHK03~\cite{li2014deepreid} and MSMT17~\cite{wei2018person}) are divided into two parts: three datasets as the source domains for training and the remaining one as the target domain for testing. For simplicity, in the following sections, we denote Market1501 as M, DukeMTMC as D, CUHK02 as C2, CUHK03 as C3, MSMT17 as MT and CUHK-SYSU as CS. The two different evaluation settings are shown in Tab.~\ref{table:Protocol}.

\textbf{Evaluation metrics:} We follow the common evaluation metrics person Re-ID, \textit{i.e.}, mean Average Precision (mAP) and Cumulative Matching Characteristic (CMC) at Rank-$k$.

\subsection{Implementation Details}

We adopt ResNet50~\cite{he2016deep} pretrained on Imagenet as our backbone. Following previous method~\cite{luo2019strong}, the last residual layer’s stride size is set to 1. A generalized mean Poolineg (GeM)~\cite{radenovic2018fine} with a batch normalization layer is used after the backbone to obtain the Re-ID features. Images are resized to 384 $\times$ 128, and the training batch size is set to 128, including 8 identities and 16 images per identity. For data augmentation, we use random flipping, random cropping and color jittering. 
We train the model for 60 epochs. The learning rate is initialized as $3.5\times10^{-4}$ and divided by 10 at 40th epochs, weight dacay is $5\times10^{-4}$. $\lambda$ in Eq.~\ref{eq:sample} is set to 0.6.
For our Baseline, we combine all the source domains into a hybrid dataset, and employ the same settings for the loss function, data augmentation strategy and the backbone, and then train the model without GDNorm. To speed up the training process and increase memory efficiency, we use automatic mixed-precision training~\cite{micikevicius2017mixed} in the entire process. All experiments are conducted on a single NVIDIA Titan XP GPU.

\subsection{Comparison to state-of-the-art methods}



\textbf{Comparison under the Protocol-1.} As shown in Tab.~\ref{table:result1}, `MS' is the multiple source datasets under Protocol-1 (M, D, C2, C3 and CS). `Comb1' is the leave-one-out setting for VIPeR, PRID, CUHK01, iLIDS and CAVIAR datasets. `Comb2' is the combination of C3, D and synthetic datasets. 

The comparison methods are mainly divided into three groups: supervised learning (S), unsupervised domain adaptation (UDA), and domain generalization (DG). Supervised methods suffer from the over-fitting issue due to the extremely small scale of training set. In contrast, our GDNorm method achieves better performance than all the supervised methods without access to labeled target data. Furthermore, we can observe that even if unlabeled target samples can be utilized for UDA methods, our GDNorm is still superior to these UDA methods without using any target data. Moreover, the mean performance on four target domains of our GDNorm method improves RaMoE~\cite{dai2021generalizable} by 7.4\% R-1 accuracy and 7.2\% mAP, and DIR-ReID~\cite{zhang2021learning} 6.0\% R-1 accuracy and 6.0\% mAP, respectively. Previous DG person Re-ID methods directly adopt the model trained on the source domains to unseen target domain without considering the biased issue. Compared to these methods, our proposed method finds the debiased feature statistics from an accurate estimated Gaussian process and significantly enhances the generalization ability.

\textbf{Comparison under the Protocol-2.} As shown in Tab.~\ref{table:result2}, we compare the proposed GDNorm with $QAConv_{50}$~\cite{liao2019interpretable}, $M^3L$~\cite{zhao2020learning} and RaMoE~\cite{dai2021generalizable} under Protocol-2. GDNorm outperforms the performances of these methods by a large margin. Specifically, our method improves the second best RaMoE by 4.5\% R1 accuracy and 11.7\% mAP on Market-1501. When testing on DukeMTMC, GDNorm improves the second best RaMoE by 4.6\% and 6.9\% in terms of R1 accuracy and mAP. When testing on CUHK03, GDNorm outperforms RaMoE by 10.0\% R1 accuracy and 10.4\% mAP. GDNorm also improves RaMoE by 14.0\% R-1 accuracy and 6.9\% mAP on MSMT17. The performances on these four large-scale ReID datasets have demonstrated the strong domain generalization of our GDNorm through debiased BN estimation from the accurate Gaussian process.

\subsection{Ablation Study}


\textbf{Effectiveness of components in GDNorm.} To investigate the effectiveness of each component in DGNorm, we conduct ablation studies in Tab.~\ref{table:components}. `E' denotes the model directly trained with domain-specific BN layers and tested with the mean parameters of the domain specific BN layers across all source domains. `S' means employing the mean path from the learned Gaussian process as the debiased BN parameters. As shown in Tab.~\ref{table:components}, the performance of `E' outperforms the baseline by 5.2\% R1 accuracy and 8.3\% mAP. It indicates that the domain-specific BN layers can effectively capture the specific feature statistics information from the corresponding domains, which can be integrated to make full use of all the effective information of source domains for enhancing the generalization. On this basis, the model with the dynamically self-refining Gaussian process improves the performance by 2.8\% R1 accuracy and 4.7\% mAP, which demonstrates the effectiveness of the learned debiased feature statistics of BN layers, tending to real hyper-distribution.

\begin{table}[tb]
    \small
    \caption{Evaluation of the influence of different paths as well as the inference time under Protocol-2.}
	\centering
	\vspace{-0.20cm}
	\begin{tabular}{c|ccccc}
		\Xhline{3\arrayrulewidth}
		\multirow{2}{*}{Testing Paths} & \multicolumn{5}{c}{D+MT+C3$\rightarrow$M} \\ \cline{2-6}
		& R1 & R5 & R10 & mAP & Time \\ \hline
		Single Path (D) & 75.5 & 89.3 & 92.1 & 48.0 & 0.09s / batch \\
		Single Path (MT) & 80.8 & 90.8 & 92.2 & 56.4 & 0.09s / batch \\
		Single Path (C3)  & 78.9 & 91.4 & 93.7 & 54.2 & 0.09s / batch \\
		\textbf{Mean Path}& \textbf{86.5} & \textbf{95.2} & \textbf{96.9} & \textbf{68.2} & 0.09s / batch \\
		
		\Xhline{3\arrayrulewidth}
	\end{tabular}
	\label{tab:paths}
\end{table}

\begin{table}[tb]
    \small
	\caption{Ablation study on sample ratio $\lambda$ under Protocol-2.}  
	\vspace{-0.20cm}
	\centering
	\begin{tabular}{c|cccc}
		\Xhline{3\arrayrulewidth}
		\multirow{2}{*}{$\lambda$} & \multicolumn{4}{c}{D+MT+C3$\rightarrow$M} \\ \cline{2-5}
		 & R1 & R5 & R10 & mAP  \\ \hline
		0.1  &  83.9 & 93.6 & 95.5 & 63.2  \\ \hline
		0.5  &  86.3  &  94.4  &  96.2  &  67.7  \\
		0.6  & \textbf{86.5} & \textbf{95.2} & \textbf{96.9} & \textbf{68.2} \\ 
		0.7  &  85.8 & 94.2 & 95.9 & 67.3\\ \hline
		1    &  84.3 & 92.8 & 93.2 & 64.4 \\
		\Xhline{3\arrayrulewidth}
	\end{tabular}
	\vspace{-0.50cm}
	\label{tab:sample ratio}
\end{table}

\begin{figure}[!t]
	\begin{center}
		\includegraphics[width=0.98\linewidth]{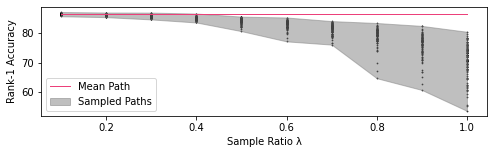}
	\end{center}
	\vspace{-0.20cm}
	\caption{Performance comparison between the mean path and other sampled paths under Protocol-2.}
	\vspace{-0.50cm}
	\label{fig:visual}
\end{figure}

\textbf{Analysis of the influence of different paths.} We analyze the performance of using different paths from the estimated Gaussian process as the optimal debias parameters of BN layers for the target domain. As shown in Fig.~\ref{fig:visual}, we randomly sample paths from the learned Gaussian process to serve as BN layers in the model and test their performance. We change $\lambda$ in Eq.~\ref{eq:sample} from 0.1 to 1, and randomly sample 100 paths at each $\lambda$ as 100 different models. Their results are marked with black dots. The grey region is the area between the best sampled path's model and the worst sampled path's model at each $\lambda$. The horizontal line in red represents the performance of the ``Mean Path''. From the results in the Fig.~\ref{fig:visual}, few number of models generated by randomly sampled paths perform slightly better than the ``Mean Path'', which indicates that these models more closely match the target domain. But from the perspective of average risk, ``Mean Path'' achieves the best performance. Moreover, as shown in Tab.~\ref{tab:paths}, ``Single Path (D)/(MT)/(C3)'' means inference with the parameters of domain-specific BN layers learned from D/MT/C3 domain-specific model, respectively. ``Mean Path'' means to directly use the mean path of the Gaussian process to serve as every BN layer in  the  model. ``Single Path (D)/(MT)/(C3)''  are inferior to ``Mean Path''. It indicates that our GDNorm makes use of all the domain-specific models' features to improve the generalization ability. GDNorm only takes 0.09 seconds per query batch images to inference, which is just one-third time cost of the ensemble model with three domain-specific networks, showing the high efficiency of our GDNorm.

\textbf{Effectiveness of the sampling ratio $\lambda$.} The results in Tab.~\ref{tab:sample ratio} show the influence of different sampling ratio $\lambda$ in Eq.~\ref{eq:sample} during the training stage. As shown in Tab.~\ref{tab:sample ratio}, we can observe that when $\lambda$ increases from 0.1 to 0.6, the proposed method obtains 2.4\% and 5.0\% improvements in terms of R-1 accuracy and mAP, respectively. When $n_d$ increases from 0.6 to 1, the performance decreases by 2.2\% and 3.8\% R-1 in terms of accuracy and mAP, respectively. GDNorm obtains the best performance when $\lambda$ is set to 0.6.


\section{Conclusion}
In this paper, we propose a novel Debiased Batch Normalization via Gaussian Process approach (GDNorm) for generalizable person re-identification. GDNorm models the feature statistic estimation (the parameters) from the domain-specific BN layers as a dynamically self-refining Gaussian process to fit the real hyper-distribution of domains for improving the generalization ability on unseen target domain. It employs the self-refining mechanism to further optimize these learned parameters of domain-specific BN layers for source domains, and estimates more accurate Gaussian process by them in return, tending to real data distribution. In addition, GDNorm directly exploits the mean path of the Gaussian process as the optimal BN estimating for unseen target domain, without requiring no extra computational cost as compared to ensemble learning. Extensive experiments show the proposed GDNorm achieves the state-of-the-art performance on multiple standard benchmarks. 

\section{Acknowledgments}
This work was supported by the National Key R\&D Program of China under Grant 2020AAA0105702, National Natural Science Foundation of China (NSFC) under Grant U19B2038 and 62106245, the University Synergy Innovation Program of Anhui Province under Grants GXXT-2019-025, and the Fundamental Research Funds for the Central Universities under Grant WK2100000021.

\bibliography{aaai22}

\end{document}